%% file: main.tex
\documentclass[5p,times,procedia]{elsarticle}

\usepackage{ecrc}
\usepackage{algorithm}
\usepackage{algorithmic}
\usepackage{graphicx}%插入图片
\usepackage{amssymb}%数学符号
\usepackage{amsthm}%数学定理
\usepackage{amsmath}%数学公式、矩阵、积分求和等
\usepackage{lineno}%显示行号
\usepackage{txfonts} %默认字体times new roman
\usepackage{enumitem}%项目编号
\usepackage{multirow} %多行合并
\usepackage{caption} %改变图表标题
\usepackage{txfonts} %默认字体times new roman
\usepackage{array} %\调用公式宏包的命令应放在调用定理宏包命令之前，也能控制表格
\usepackage{booktabs} %调整表格线与上下内容的间隔
\usepackage{longtable}%调用跨页表格
\usepackage{bm}%数学字体加粗
\usepackage{setspace} %调整一段文字的行间距
\usepackage{natbib} %参考文献管理包

\volume{00}

\firstpage{1}

\journalname{}

\runauth{}

\jid{procs}

\jnltitlelogo{}

\usepackage{amssymb}
\biboptions{sort&compress}%参考文献可以压缩显示例如1-3
\allowdisplaybreaks[4] %允许公式跨页
\usepackage[figuresright]{rotating}

\begin{document}

\begin{frontmatter}

%% Title, authors and addresses

%% use the tnoteref command within \title for footnotes;
%% use the tnotetext command for the associated footnote;
%% use the fnref command within \author or \address for footnotes;
%% use the fntext command for the associated footnote;
%% use the corref command within \author for corresponding author footnotes;
%% use the cortext command for the associated footnote;
%% use the ead command for the email address,
%% and the form \ead[url] for the home page:
%%
%% \title{Title\tnoteref{label1}}
%% \tnotetext[label1]{}
%% \author{Name\corref{cor1}\fnref{label2}}
%% \ead{email address}
%% \ead[url]{home page}
%% \fntext[label2]{}
%% \cortext[cor1]{}
%% \address{Address\fnref{label3}}
%% \fntext[label3]{}

% \dochead{}
%% Use \dochead if there is an article header, e.g. \dochead{Short communication}
%% \dochead can also be used to include a conference title, if directed by the editors
%% e.g. \dochead{17th International Conference on Dynamical Processes in Excited States of Solids}

\title{Towards Accurate Deceptive Opinions Detection based on Word Order-preserving CNN}

%% use optional labels to link authors explicitly to addresses:
%% \author[label1,label2]{<author name>}
%% \address[label1]{<address>}
%% \address[label2]{<address>}

% \author[label1]{Siyuan Zhao}
% \author[label1,label2]{Zhiwei Xu}
% \author[label1]{Limin Liu}
% \author[label1]{Mengjie Guo}
% \small Email: \{siyuan\_zhao\}@foxmail.com,\{xuzhiwei2001\}@ict.ac.cn,\{limin\_ll\}@foxmail.com,\{mengjie\_guo\}@foxmail.com
% }
% \address[label1]{College of Information Engineering, Inner Mongolia University of Technology, Hohhot, Inner Mongolia, China}
% \address[label2]{University of Chinese Academy of Sciences, Beijing, China}
\author[First]{Siyuan Zhao}
\ead{siyuan\_zhao@foxmail.com}
\ead[url]{www.foxmail.com}

\author[Second]{Zhiwei Xu\corref{cor1}}
\ead{xuzhiwei2001@ict.ac.cn}
\ead[url]{www.ict.ac.cn}

\author[First]{Limin Liu}
\ead{limin\_ll@foxmail.com}

\author[First]{Mengjie Guo}
\ead{mengjie\_guo@foxmail.com}

\author[First]{Jing Yun}
\ead{yunjing\_zoe@163.com}

\cortext[cor1]{Corresponding author at:University of Chinese Academy of Sciences, Beijing, China.}

\address[First]{College of Information Engineering, Inner Mongolia University of Technology, Hohhot, Inner Mongolia, China. }
\address[Second]{University of Chinese Academy of Sciences, Beijing, China.}

\begin{abstract}
% * <xuzhiwei2001@ict.ac.cn> 2018-01-28T16:18:13.780Z:
% 
% 和introduction前两段类似，说深度学习已经广泛应用，在自然语言学习方面，鉴于其高度的灵活性，已经实现了对复杂语义的分析。虚假评论识别是深度学习模型的一个重要应用领域，相关机制已经得到了重视和研究。基于深度学习的识别机制具有更好的自适应性，可以有效识别各类虚假评论。
% 
% ^.

Nowadays, deep learning has been widely used. In natural language learning, the analysis of complex semantics has been achieved because of its high degree of flexibility. The deceptive opinions detection is an important application area in deep learning model, and related mechanisms have been given attention and researched. On-line opinions are quite short, varied types and content. In order to effectively identify deceptive opinions, we need to comprehensively study the characteristics of deceptive opinions, and explore novel characteristics besides the textual semantics and emotional polarity that have been widely used in text analysis. The detection mechanism based on deep learning has better self-adaptability and can effectively identify all kinds of deceptive opinions.
In this paper,  we optimize the convolution neural network model by embedding the word order characteristics in its convolution layer and pooling layer, which makes convolution neural network more suitable for various text classification and deceptive opinions detection. The TensorFlow-based experiments demonstrate that the  detection mechanism proposed in this paper achieve more accurate deceptive opinion detection results.
\end{abstract}

\begin{keyword}
%% keywords here, in the form: keyword \sep keyword
Natural language processing \sep deceptive opinion spam detection \sep deep learning \sep convolution neural network \sep text classification 
%% PACS codes here, in the form: \PACS code \sep code

%% MSC codes here, in the form: \MSC code \sep code
%% or \MSC[2008] code \sep code (2000 is the default)

\end{keyword}

\end{frontmatter}

%%
%% Start line numbering here if you want
%%
% \linenumbers

%% main text

\input{1introduction}

\input{2background}

\input{3model}

\input{4scheme}

\input{5analysis}

\section{Acknowledgment}
This work was supported by Natural Science Foundation of China (61540004, 61502255 and 61650205), and was supported by Natural Science Foundation of Inner Mongolia Autonomous Region (2017MS(LH)0601).

%% The Appendices part is started with the command \appendix;
%% appendix sections are then done as normal sections
%% \appendix

%% \section{}
%% \label{}

%% References
%%
%% Following citation commands can be used in the body text:
%% Usage of \cite is as follows:
%%   \cite{key}         ==>>  [#]
%%   \cite[chap. 2]{key} ==>> [#, chap. 2]
%%

%% References with BibTeX database:

\bibliographystyle{elsarticle-num}
% \bibliography{myreference.bib}

%% Authors are advised to use a BibTeX database file for their reference list.
%% The provided style file elsarticle-num.bst formats references in the required Procedia style

%% For references without a BibTeX database:

% \begin{thebibliography}{00}

%% \bibitem must have the following form:
%%   \bibitem{key}...
%%

% \bibitem{}

% \end{thebibliography}

\end{document}

%% file: 1introduction.tex
\section{Introduction}
\label{sec:intro}

Artificial neural networks (ANNs) is a well known bio-inspired model that simulates human brain capabilities such as learning and generalization~\cite{Black2001Method,Pulverm2009Discrete}. ANNs consist of a number of interconnected processing units, wherein each unit performs a weighted sum followed by the evaluation of a given activation function~\cite{Knoblauch2013Neural}. ANNs has the ability of self-learning, associative storage capabilities and high-speed search for optimal solution~\cite{D2002Behavioral}. In recent years, ANNs has been applied in natural language processing, pattern recognition, knowledge engineering, expert systems, .etc. 

% Artificial Neural Networks (ANNs) is abstract and simulation of several basic characteristics of the human brain or natural neural network~\cite{Black2001Method,Pulverm2009Discrete}. Artificial neural networks is based on physiological research on the brain and aim at simulating certain mechanisms of the brain~\cite{Knoblauch2013Neural}. Artificial neural network has the ability of self-learning, associative storage capabilities and high-speed search for optimal solution~\cite{D2002Behavioral}. Therefore, artificial neural networks have been applied in natural language processing, pattern recognition,  knowledge engineering, expert systems, .etc. 
The concept of deep learning originates from the study of artificial neural networks. Multi-layer perceptron with multiple hidden layers is a deep learning model. The deep learning can achieve the feature selection and organization of high dimensional data, and update the model parameters dynamically according to the feedback. It can adaptively detect deceptive opinions. 
% * <xuzhiwei2001@ict.ac.cn> 2018-01-28T07:57:38.538Z:
% 
% 神经网络和深度学习不完全一样，你得引导到或者直接进入深度学习模型，然后说深度学习模型在自然语言处理上由于具有什么特点因此得到了广泛使用
% 
% ^ <xuzhiwei2001@ict.ac.cn> 2018-01-28T14:16:48.803Z:
% 
% 抓紧改第一段
%
% ^.

Deceptive opinions detection is an important application of deep learning model. 
The existence of deceptive opinions makes customers who are lack of relevant experience difficult to obtain accurate judgments of the reviewed products and buy the appropriate products. To achieve effective deceptive opinions detection, representative training data sets are highly desired. There are two types of training data sets, the constructed the data sets based on the semantic or  polarity analysis of on-line opinions[9], the true  data sets of user opinions[25,26,28], which include opinion texts[13,14,15,16],  behaviors of users or between users[10,11,16]. 
In addition, the inputed options are classified by support vector machine[12] and other machine learning methods.  On-line opinions are short texts, varied in types and content, these  existing approaches cannot adapt to various short texts and detect deceptive opinions with high accuracy. In order to achieve  effective deceptive opinions identification, we need to adapt all  relevant features of deceptive opinions to design a comprehensive deep learning model of deceptive opinions identification. 

Considering the sparse and various expression of text opinions, we first introduce the text word order into the process of the deceptive opinion analysis. In this way, we expand the characteristic dimension of our deceptive opinions model and proposes a novel word order-preserving pooling layer,  which is additionally embedded in the existing CNN (Convolutional Neural Network) model to improve the deceptive opinions detection effectively.

\noindent\textbf{Contributions.} The main contributions are as follows:
\begin{enumerate}
\item Since on-line opinions are short and various in forms, this paper introduces a novel feature of opinion texts, the word order of opinions, and proposes a word order-preserving CNN model, which preserves the word order characteristics in the process of opinion text feature analysis. 
\item We implement our word order-preserving CNN (OPCNN) model on an open source deep learning platform, TensorFlow, and demonstrate that compared with basic CNN model, OPCNN can achieve more accurate detection for deceptive opinions.
\end{enumerate}

\noindent\textbf{Organization}. In section~\ref{sec:related}, we introduce the related work. Section~\ref{sec:model} gives details of our proposed deep learning model, and Section~\ref{sec:scheme} provides the performance evaluation based on TensorFlow. We conclude in Section~\ref{analysis}.
%This section is organized as follows,

%% file: 2background.tex
\section{Background and Related Work}
\label{sec:related}

\subsection{CNN Model and Its Applications}
The neural network model is connected with a large number of neurons to form a complex network system with adaptive and self-learning ability, and suitable for dealing with the unclear inherent characteristics of the data. As a new class of neural network model, the deep learning model can be used to learn the characteristics of various real things from large-scale data sets, and these features can be directly applied to various computing models by the computer. 

CNN (Convolutional Neural Network), is a type of deep learning models and a research hotspot in recent years. CNN has a good fault tolerance, parallel processing and self-learning ability~\cite{Wei2013Study} and is widely used in image processing, speech recognition, natural language processing and other fields, and has been widely used in the text classification. Compared with other popular neural networks RNN~\cite{2015rnn}(Recurrent Neural Networks), the results of the analysis in the field of text classification are similar. Moreover, due to the opinion is generally a short sentence text, convolution function of the overall structure of the sentence has a general ability, which makes CNN in dealing with short text when the accuracy rate slightly better. Compared with the RNN, CNN's training time is shorter, more efficient, to save time costs. Due to the short length of the deceptive opinions, compact structure, and independently expressing the meaning of the characteristics of short text analysis task, it is possible for CNN to deal with deceptive opinions detection.

\subsection{Related Work}
Several researches on deceptive opinions detection have been proposed. Jindal and Liu~\cite{Jindal2008Opinion} first studied deceptive opinions problem and trained models using features based on the opinion content, user, and the product itself. Myle Ott et al.~\cite{ott2011finding} created a benchmark dataset by employing Turkers to write fake opinions. Fei et al.~\cite{Fei2013Exploiting} proposed that a large number of opinions made use of a sudden burst either caused by the sudden popularity of the product or by a sudden invasion of a large number of fake opinions, including some of the features of real users. Markov Random Field (MRF) was used to construct users and their co-occurrence in emergencies by establishing a network for critics in different periods of emergency. Finally, Belief Propagation(BP) was used to infer whether a user is a fake user or not.Wang et al.~\cite{Wang2015Semantic} proposed an innovative heterogeneous opinion graph model to capture the relationship between the users and users' opinions on the shop, and used the interaction and the role of the nodes in the figure to reveal the causes of deceptive opinion, and designed an iterative algorithm to identify deceptive opinions. Mukherjee~\cite{Mukherjee2013What} et al. found that more than 70\% of deceptive opinion publishers issued opinions between the similarity is greater than 0.3, and real opinion publishers published opinions similarity between less than 0.18 in the Yelp data set. The content similarity calculation for the opinions made by the same commentator can reflect the characteristics of the opinion's behavior.

There have been some studies using deep learning models to identify deceptive opinions. Raymond~\cite{Lau2012Text} team builds a semantic language model to identify semantic repetitive opinions and makes deceptive opinions detection. However, due to the opinion itself has a certain degree of semantic similarity and content on the repeatability, there may be a miscarriage of justice. Li et al.~\cite{Li2015Learning} took the word vector as input, with CNN, the emotional polarity feature can also be applied to unsupervised methods for deceptive opinions text detection. However, only considering the emotional polarity of the deceptive opinion on the identification is not sufficient. At the same time, the local sampling of the CNN model can not take into account the existence of the word order in the text. Jindal~\cite{Jindal2010Finding} thought that the same user that gives his all positive opinions or negative opinions to the same brand of products is a kind of abnormal behavior and the corresponding opinion maybe deceptive opinion. The researchers proposed a "one-condition rules" and a "two-condition rules" model, to predict the falseness of the text by probabilistic prediction. Yapeng Jing~\cite{Jingyapeng2014} sets the data set on the AMT of hotel opinions, uses the information gain to select the feature of the word bag and then detects deceptive opinions through the ordinary neural network, DBN-DNN network and LBP network. However, the artificial data set can not accurately reflect the true opinions.

Deceptive opinions detection is a type of complex text classification. The deceptive opinion is very short, varied type and content. In order to effectively identify deceptive opinions, besides the textual semantics and emotional polarity that have been widely used in text analysis, we need to further extract the deep features of deceptive opinions to characterize deceptive opinion effectively. Therefore, we introduce the word order into the CNN model, design the preservation of the k-max pooling technology and expand the deceptive opinion feature mining range to solve the difficulties in the identification of deceptive opinions and to enhance the accuracy of deceptive opinions detection.

%% file: 3model.tex
\section{Deceptive Opinion Detection Model}
\label{sec:model}
To achieve an accurate deceptive opinion detection, we study the word order characteristic in on-line opinion texts. In addition, we design a word order-preserving CNN network to model various short opinion texts. In this way, we embed a foundational textual characteristic into deceptive opinion detection process and obtain more accurate detection results.

\begin{figure}
\centering
\vspace{-100pt}
  \includegraphics[width=8cm]{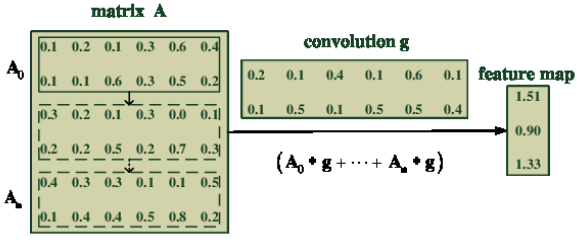}\\
    \vspace{-100pt}
  \caption{The process of convolution in the CNN model deals with the data.  The matrix A is multiplied by the corresponding elements of the convolution matrix g and summed up. During the process of convolution, we will get the feature map.}\label{fig:cnn}
\end{figure}

\subsection{Chinese Word Order}

Almost all languages have its word ordering of the subject(S), object(O), and the verb(V), and among the languages of the world, all six possible basic word orders exist~\cite{Dryer2005Order} especially SVO((Subject-Verb-Object) and SOV(Subject-Object-Verb). The study has shown that the earliest human language had rigid word order. Nowadays, SOV basic word order is common among the languages of the world and that many other word orders can be reconstructed back to an SOV stage. It can be concluded that SOV must have been the word order of the 'ancestral language' among the six possible word orders~\cite{FJ2000On,Gellmann2011The}. In addition, there are researches to demonstrate that besides SOV, SVO is such a prominent word order in the languages of the world. For example, a sentence like 'fireman kicks boy', both nouns could in principle be the agent. SVO is used to avoid expressing two plausible agents ('fireman' and 'boy') at the same side of the verb instead of SOV~\cite{Gibson2013A}. As a traditional language, Chinese text also possesses word order(SVO). Word order in Chinese text is an inherent feature of text classification. In this paper, in order to describe the short opinion text, we need the word order feature to the process of detective opinion feature mining, and optimize CNN model to identify deceptive opinions. Moreover, we will use the sentence with word order as the input of our model to prove the idea of word order-preserving in this paper.

% \begin{figure}[t]
% \setlength{\abovecaptionskip}{0.cm}
% \setlength{\belowcaptionskip}{-2.cm}
%   \centering
%   \includegraphics[height=86mm,width=91mm]{KMNN.eps}
%   \caption{ A OpCNN for the nine word input sentence. Word embeddings have size d = 5. The network has one convolution layer with three feature maps each. The widths of the each filter at the layer are 3, 4 and 5. The k-max pooling layers have values k of 3.}\label{fig:KMNN}
% \end{figure}

\begin{figure}
\centering
  \includegraphics[width=9cm]{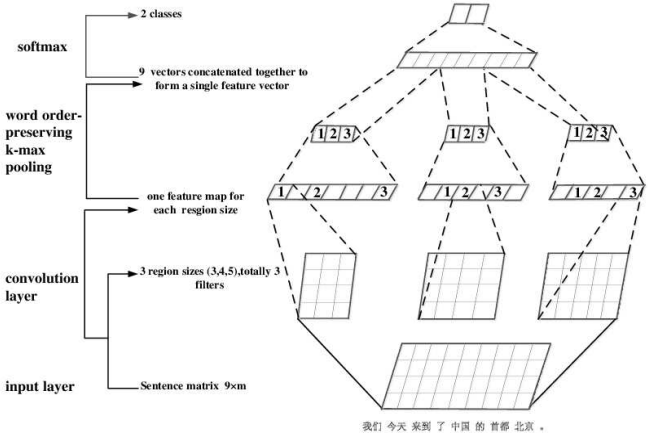}\\
  \caption{OPCNN for the nine word input sentence.The network has one convolution layer with three feature maps each. The width of the each filter at the layer are 3, 4 and 5.}\label{fig:KMNN}
\end{figure}

\subsection{OPCNN model}

CNN model includes input layer, convolution layer, pooling layer and output layer. We proposed an improved CNN model considering the Chinese word order characteristic. The input layer takes the opinion sentences with a certain word order as input values. In convolution layer and the pooling layer, we preserve the word order of inputed sentences and apply the word order persevering pooling method instead of the original pooling layer,as shown in Fig.1. Ultimately, optimize the characteristic selection process of CNN model is optimized (the detailed model is illustrated in Fig.2). 

\subsubsection{Input Layer}
We use word vectors to represent the word frequency of each word~\cite{Johnson2014Effective} and take them as the training inputs of our model. We use the word2vec model to predict words that appear in the context by training a neural network language model to generate word vectors. The input layer consists of an ${n \times m}$ two-dimensional matrix, where n is the length of the sentence and m is the dimension of word vectors. The text representation process can be formulated as Eq.1, where a represents the matrix, w represents word vector of every word and v represents the value of every word vector. Ultimately, each opinion is represented by a two-dimensional word vector matrix. 

\begin{normalsize}
 \begin{eqnarray}
\left\{ \begin{array}{l}
A = {\left( {{{\vec w}_1},{{\vec w}_2},...,{{\vec w}_n}} \right)^T}\\
{{\vec w}_i} = \left( {{v_1},{v_2},...,{v_m}} \right)
\end{array} \right.
 \end{eqnarray}
 \end{normalsize}
 
\subsubsection{Convolution Layer}

The input layer transfers the word vector matrix A to the convolution layer for convolution operations. The padding of convolution has two types:same and valid. As is shown in Eq.2, we perform the i th convolution in the l-layer ${K_i^l}$ on matrix A, taking the ReLU function as activation function, the bias ${b_i^l}$ as the the valid padding of convolution, and the matrix ${a_i^l}$ as feature map. The size of the convolution window is ${h \times m}$, where h is the width of the convolution kernel and m is the dimension of the word vector. The width of the convolution kernel(h) needs to be set and adjusted dynamically, as is shown in Fig.1. As the convolution kernel continues to move down, the corresponding eigenvalues of the convolution kernel are generated. According to this convolution window, we will get a few of all ``1'' columns on the feature map.
% * <xuzhiwei2001@ict.ac.cn> 2018-01-28T14:44:04.756Z:
% 
% 后面我删了，看不懂，我只能顺到这，你自己看看，接着写这一层最终得到了什么
% 
% ^.

\begin{normalsize}
 \begin{eqnarray}
a_i^l = \sigma ( conv2(A,K_i^l,valid) + b_i^l)
\end{eqnarray}
 \end{normalsize}
 
%  \begin{normalsize}
%  \begin{eqnarray}
% u_i^l{\rm{ = }}C_i^l
% \end{eqnarray}
%  \end{normalsize}
 
%  \begin{normalsize}
%  \begin{eqnarray}
% a_i^l = \sigma (u_i^l)
% \end{eqnarray}
%  \end{normalsize}
 
The input value of the window is converted to an eigenvalue by the nonlinear transformation of the neural network. As the window moves down, the corresponding eigenvalues of the convolution kernel are generated and the eigenvectors corresponding to the convolution kernel are formed. We use the nonlinear transformation activation function called ReLU.
 
\subsubsection{Word Order Persevering Pooling Layer}

The word order persevering pooling layer reduces the number of feature parameters. The output of the order pooling layer is the maximum value of each feature map. The max pooling method can keep the location of the feature and the invariance value of the pooling operation. This feature affects the accuracy of text analysis, since the Chinese texts exists the word order characteristics. The position of each word in a sentence is a very important feature in the text analysis, so it is particularly important to preserve the word order of the sentences. Thus, the word order persevering k-max pooling method is proposed here to replace the original max pooling method in the paper.

\begin{normalsize}
 \begin{eqnarray}
a_i^l =\sigma( \beta _i^l \times pooling(a_i^{l - 1},k) + b_i^l) 
\end{eqnarray}
\end{normalsize}

% \begin{normalsize}
%  \begin{eqnarray}
% u_i^l{\rm{ = }}S_i^l
% \end{eqnarray}
% \end{normalsize}

% \begin{normalsize}
%  \begin{eqnarray}
% a_i^l{\rm{ = }}\sigma \left( {u_i^l} \right)
% \end{eqnarray}
% \end{normalsize}
 
As is shown in Eq.3, we use the word order persevering k-max pooling method generally to deal with the result ${a_i^{l - 1}}$ of the convolution (l-1)layer. The method idea is to select the k maximum values from the one-dimensional feature map obtained from the previous convolution layer operation, and discard the other eigenvalues.

As shown in Fig.3, the word order-preserving k-max pooling method selects the k highest values in the sequence s,  where length of s is longer than k. The order of the selected values corresponds to their original order in s. 
% * <xuzhiwei2001@ict.ac.cn> 2018-01-28T15:07:39.005Z:
% 
% 下面这一段看不懂，自己用简单的语言写清楚，还是和前面一样，最后落到你这层得到了什么
% 
% ^.
The word order-preserving k-max pooling method can discern more finely the number of times that the feature is highly activated in s~\cite{Kalchbrenner2014A} than that of max-pooling methods. What is more, the method can also distinguish the progression by which the high activations of the feature change across s. In this method, we can get the k highest feature values in the sequence s.

\begin{figure}
\centering
\vspace{-70pt}
  \includegraphics[width=0.4\textwidth]{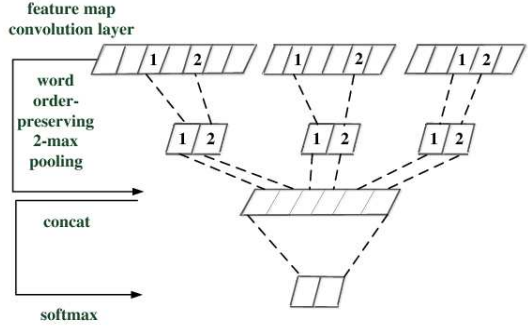}\\
  \vspace{-100pt}
  \caption{The method in the pooling layer in our paper.We let the k equal to 2, and in the convolution layer the model has three types convolution.}
\label{fig:pooling_method}
\end{figure}

\subsubsection{Output Layer}

We concat the obtained features form the pooling layer. It is a two classification problem which distinguishes the deceptive opinion from real opinion. The result of the concat function is then entered into the softmax function to assess the probability that the opinion is deceptive. 
% * <xuzhiwei2001@ict.ac.cn> 2018-01-28T15:11:50.707Z:
% 
% 补全
% 
% ^.
Finally, we use cross entropy as a model of the loss function to measure the difference between the predicted value and the true value in the OPCNN model.

\begin{algorithm}[t]
\begin{small}
\caption{Forward Propagation Algorithm} %算法的名字
\hspace*{0.02in} {\bf Input:train dataset ${{\rm{\{ (}}{{\rm{x}}_n}{\rm{,}}{{\rm{t}}_n}{\rm{)\} }}_{n = 1}^N}$,the structure of CNN ${\left\{ {{\rm{ }}{h_l}} \right\}_{l = 1}^L}$} %算法的输入， \hspace*{0.02in}用来控制位置，同时利用 \\ 进行换行

\hspace*{0.02in} {\bf Output:The parameters of CNN} %算法的结果输出

\begin{algorithmic}[1]
\FOR{$i < L$} % If 语句，需要和EndIf对应
  \IF{${h_l}$ is convolution layer Algorithm}
    \FOR{$\forall i$}
     \STATE  ${a_i^l = \sigma (conv2(A,K_i^1,valid) + b_i^l)}$
    \ENDFOR
  \ENDIF
  \IF{${h_l}$ is pooling layer}
    \FOR{$\forall i$}
    \STATE  ${a_i^l = \sigma (\beta _i^l \times pooling(a_i^{l - 1},k) + b_i^l)}$
    \ENDFOR
  \ENDIF
  \IF{${h_l}$ is connection layer}
    \STATE  ${a_i^l = \sigma ({z^l})}$
  \ENDIF
\ENDFOR
\end{algorithmic}
\end{small}

\end{algorithm}

\begin{algorithm}[t]
\begin{small}
\caption{Back Propagation Algorithm} %算法的名字

\hspace*{0.02in} {\bf Input:The parameters of CNN,train dataset ${{\rm{\{ (}}{{\rm{x}}_n}{\rm{,}}{{\rm{t}}_n}{\rm{)\} }}_{n = 1}^N}$,the structure of CNN ${\left\{ {{\rm{ }}{h_l}} \right\}_{l = 1}^L}$} %算法的输入， \hspace*{0.02in}用来控制位置，同时利用 \\ 进行换行

\hspace*{0.02in} {\bf Output:Response error} %算法的结果输出

\begin{algorithmic}[1]
\FOR{$i < L$} % If 语句，需要和EndIf对应
  \IF{${h_l}$ is convolution layer}
    \FOR{$\forall i,j$}
     \STATE  ${\frac{{\partial J}}{{\partial K_{ij}^l}} = \sum\limits_{st} {{{\left( {{\delta ^{(l)}}} \right)}_{st}}} {\left( {P_j^{(l - 1)}} \right)_{st}}}$
    \ENDFOR
    \FOR{$\forall i$}
     \STATE  ${\frac{{\partial J}}{{\partial b_i^{(l)}}} = \sum\limits_{st} {{{\left( {{\delta _i}} \right)}_{st}}}}$
    \ENDFOR
  \ENDIF
  \IF{${h_l}$ is pooling layer}
    \FOR{$\forall i$}
    \STATE  ${\frac{{\partial J}}{{\partial b_i^{(l)}}} = \sum\limits_{st} {{{\left( {{\delta _i}} \right)}_{st}}}}$
    \ENDFOR
    \FOR{$\forall i$}
    \STATE  ${\frac{{\partial J}}{{\partial \beta _i^{(l)}}} = \sum\limits_{s,t} {{{(\delta _i^{(l)} \circ d_i^{(l - 1)})}_{st}}}}$
    \ENDFOR
  \ENDIF
  \IF{${h_l}$ is connection layer}
    \STATE  ${\frac{{\partial J}}{{\partial {w^{(l)}}}} = {\delta ^{(l)}}{({a^{(l - 1)}})^T}}$
    \STATE  ${\frac{{\partial J}}{{\partial {b^{(l)}}}} = {\delta ^{(l)}}}$
  \ENDIF
\ENDFOR
\end{algorithmic}
\end{small}

\end{algorithm}

\subsection{Deceptive Opinions Detection Algorithm}

To detect deceptive opinions with OPCNN model, we manually annotate the opinion data obtained from the websites. We construct the word vector model and preprocess the experimental data. Additionally, we take the OPCNN to obtain the final text classification results, to distinguish deceptive opinions from other opinions. The complete process of deceptive opinions detection is depicted in Algorithm 1 and Algorithm 2. Specifically, we train the OPCNN model according to Algorithm 1, and detect deceptive opinions with Algorithm 2.

\textbf{Complexity analysis.} Assuming that the number of iterations is k times, the number of samples per input sentence of OPCNN model is m, the number of words of each sentence is v, the word vector dimension is d, the convolution window size is w, and the number of output channels is n. The model tackles an inputed sentence with a time complexity of O(v*n*(2d*$w^2$+w-1)). Therefore, the time complexity of the OPCNN model can be expressed as O($w^2$*k*m*n*d*v), when the model performs k iterations.

%% file: 4scheme.tex
\section{Experiments}
\label{sec:scheme}

\subsection{Experimental Data Set}
To evaluate the performance of our deceptive opinions detection scheme, we use the Ott~\cite{ott2011finding} data set. Additionally, we labeled on-line opinions including the opinions about 23,166 hotels. On some on-line opinions websites, each user can write the relevant opinions and give the evaluation level regardless of buying a product or service or not. Therefore, false reviews and false scoring phenomenons become common. To training the proposed model, we hand-annotated 10000 hotel review data by the data annotation method presented by Li~\cite{Li2011Learning}.
 
In detail, we check whether the opinion is related to the products or not, and if there is no relevance, the opinion is a deceptive opinion. In addition, we observe whether the opinion is with too much emotion, such as the opinion including a large number of commendatory with strong emotion. Similarly, opinion that contains a large number of derogatory words may also be false opinion~\cite{Crawford2015Survey,renyafeng2015}.
% * <xuzhiwei2001@ict.ac.cn> 2018-01-28T15:37:55.665Z:
% 
% 上面加个参考文献，郭孟杰当时不是说有文章是这么干的嘛
% 
% ^.

Eventually, all opinions about the 23,166 hotels are marked. Among them, 2132 opinions are fake, as depicted in Table 1. In this experiment, 80\% of the data set is used as the experimental training set, and the others are used as the experimental test set.
In order to illustrate the generalization ability of the proposed method, the data set proposed by~\cite{ott2011finding} is applied in this paper.

% \begin{table}[t]
% \begin{tabular}{c|c|c}
% \multicolumn{3}{l}{\small{\textbf{Table 1}}}\\  
% \multicolumn{3}{l}{\small{The data used in the experiment}}\\ 
% \hline 
% Comment type & Comment number & Comment length  \\ 
% \hline 
% False comment & 2132 & 78   \\ 
% \hline 
% True comment & 21034 & 112   \\ 
% \hline 
% General comment & 23166 & 109   \\ 
% \hline
% \end{tabular} 
% \end{table} 

\begin{table}[htbp]
  \caption{The data used in the experiment}
  \begin{tabular*}{\hsize}{lcc}
\hline
Opinion type & Opinion number & Opinion length  \\ 
\hline
Deceptive opinion & 2132 & 78   \\ 
Positive opinion & 21034 & 112   \\ 
All opinion & 23166 & 109   \\ 
\hline
  \end{tabular*}
\end{table}

\subsection{Implementation} 
In order to evaluate the performance of the proposed detection scheme, we implement the proposed detection scheme and three baseline schemes on TensorFlow. TensorFlow is an open source platform to implement deep learning model in piratical. 
% * <xuzhiwei2001@ict.ac.cn> 2018-01-28T16:13:39.694Z:
% 
% 下面这应该结合TensorFlow说，这是你的一个最大的亮点和贡献，但你这几乎没有基于TensorFlow的实现过程
% 
% ^.

(1) The first experimental baseline uses the classical statistical method called tf-idf for feature extraction, supports vector machine (SVM) as a classifier~\cite{asadullah2017classification} and supervises the above-mentioned tagged data.

(2) The second baseline uses Bigram to extract the feature data~\cite{Seneviratne2017Spam}. Bigram is assumed to be in a statement that the probability condition of the second word depends on one word in front of it, that is the context of a word is defined as a word that appears in front of the word~\cite{Lupker2012An}. Some of the two consecutive characters usually have the ability to represent the features of the text. Then the support vector machine (SVM) is used as the classifier to obtain the classification result.

(3) The third baseline uses the Convolution Neural Network (CNN) in the deep learning framework~\cite{huang2017detection}, combined with the short text feature extraction to apply the CNN to the deceptive comment identification. The experiment uses 3x cross validation to adjust the hyperparameters in the classifier model. The specific parameters are shown in Table 2. We use the ReLU function as a non-linear function, the super-parameter of the weight attenuation L2 is set to 0.5. Other parameters include dropout set to 0.5 and mini-batch to 50. In the CNN, we use the word2vec to get word vector as the embedding of the input layer. In the convolution layer, we use valid convolution and conv2d function to get feature map with TensorFlow. In the pooling layer, we use max pooling function to get the maximum feature value with TensorFlow. Lastly, we use the softmax function to implement test classification.

(4) We implement the OPCNN and set the parameters of OPCNN as the CNN used in the third baseline.

% \begin{table}[t]
% \begin{tabular}{c|c|c}
% \multicolumn{3}{l}{\small{\textbf{Table 2}}}\\  
% \multicolumn{3}{l}{\small{Hyperparameter setting}}\\ 
% \hline %绘制一条水平的线 
% Hyperparameter & Description & Value  \\ 
% \hline 
% d & word vector dimension & 100   \\ 
% \hline 
% $d_{min}$ & convolution width & 3,4,5   \\ 
% \hline 
% H & number of convolution & 64,64,64   \\ 
% \hline
% \end{tabular} 
% \end{table} 

\begin{table}[htbp]
  \caption{Hyperparameter setting}
   \label{extremos45}
  \begin{tabular*}{\hsize}{lcc}
\hline
Hyperparameter & Description & Value  \\ 
\hline 
d & word vector dimension & 100   \\ 

$d_{min}$ & convolution width & 3,4,5   \\ 

H & number of convolution & 64,64,64   \\ 
\hline
  \end{tabular*}
\end{table}

\subsection{Evaluation Metrics}
In order to illustrate the experimental scheme, we evaluate the experiment from five aspects: accuracy, precision, recall, f1-measure and accuracy gain.

Accuracy (A): The ratio of the samples correctly sorted by the classifier to the total number of samples for a given test data set. That is, the loss function is 0-1 loss on the test data set on the accuracy rate. true positives(TP), false positives(FP), false negatives(FN) and true negatives(TN) are the related concepts of experiment effect.

Precision (P): It calculates the ratio of all "correctly retrieved items (TP)" to all "actually retrieved (TP + FP)".

 \begin{normalsize}
 \begin{eqnarray}
P = \frac{{TP}}{{TP + FP}}
\end{eqnarray}
 \end{normalsize}
 
Recall (R): The item (TP) that is correctly retrieved is the item (TP + FN) that should be retrieved.	

 \begin{normalsize}
 \begin{eqnarray}
R = \frac{{TP}}{{TP + FN}}
\end{eqnarray}
 \end{normalsize}
 
F1-measure: F1-measure is the harmonic mean of precision and recall.
 
 \begin{normalsize}
 \begin{eqnarray}
{F_1} = \frac{{2PR}}{{P + R}} = \frac{{2TP}}{{2TP + FP + FN}}
\end{eqnarray}
 \end{normalsize}
 
Accuracy gain($\alpha$): The ratio of the experimental group method accuracy $F_e$ and the control group method $F_c$ accuracy. When the value of $\alpha$ is lager, the accuracy of the experimental group is higher than that of the control group. When the value of $\alpha$ is smaller, the accuracy of the experimental group is lower than that of the control group.

 \begin{normalsize}
 \begin{eqnarray}
\alpha {\rm{ = }}\frac{{{F_e}}}{{{F_c}}}
\end{eqnarray}
 \end{normalsize}
 
%  MSE((mean-square error):MSE is a measure that reflects the degree of discrepancy between actual value and forecast value.
 
%   \begin{normalsize}
%  \begin{eqnarray}
%   MSE=\frac{1}{n}\sum\limits_{i=1}^{n}{{{\left( {{\overset{\wedge }{\mathop{y}}\,}_{i}}-{{y}_{i}} \right)}^{2}}}
% \end{eqnarray}
%  \end{normalsize}
 
\subsection{Analysis of Results}

\subsubsection{Word Vector Dimension Selection}
In the paper, we use the word2vec model to get word vector. Firstly we should determine the dimension of the word vector because there is a certain relationship between  the word vector matrix dimension input layer and convolution kernel width in OPCNN and CNN. In this experiment, we discuss the dimension of the word vector of word2vec. The specific evaluation index is the accuracy rate, as shown in Fig.4.

\begin{figure}
\centering
  \includegraphics[width=12cm]{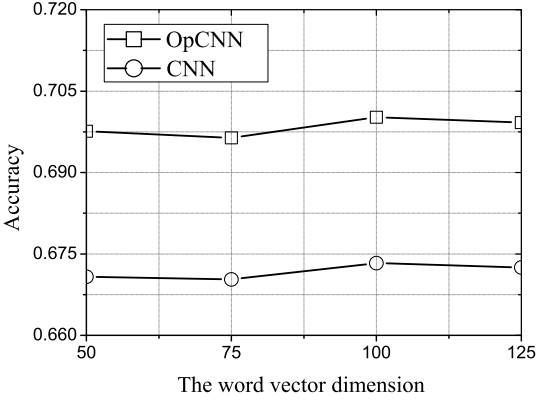}\\
  \vspace{-60pt}
  \caption{In the word2vec model, we let the dimension of the word vector 50,75,100 or 125 respectively,and use CNN and OpCNN in the experiment.}\label{fig:experiment1}
\end{figure}

It can be concluded from Fig.4 that the accuracy of the dimension 100 get better result respectively. However compared with other results, the gain of the accuracy is not obvious, and the curve shape of the diagram is not V or M. The dimension of the word vector in word2vec has the influence on the accuracy of the OPCNN and CNN moedel in our paper.

\subsubsection{K Value Selection}
In the previous chapter we mention that the k-max pooling method is used in the pooling layer instead of the original max pooling method. The essence of the k-max pooling method is the use of the top-k function, where the choice of k is particularly important. In this experiment, we discuss the effect of k value on OPCNN model. The specific evaluation index is the accuracy rate, as shown in Fig.5. In this experiment, when k is equal to 3, the accuracy reaches the maximum. This is because when the value of k is too small, it may lose the eigenvalue if model encounters the same eigenvalue. When the value of k is too large, it may get interference items to affect the accuracy.
% * <xuzhiwei2001@ict.ac.cn> 2018-01-28T15:47:06.350Z:
% 
% 这块实验有致命的问题，没有对图中结果进行描述，更没有相应的趋势的说明和原因分析，下面也有类似问题，参考这个文章的实验部分的结果说明，https://arxiv.org/abs/1511.03005
% 
% ^.
\begin{figure}
\centering
  \includegraphics[width=12cm]{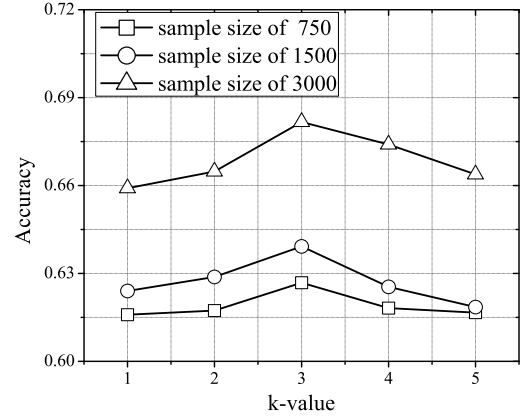}\\
  \vspace{-60pt}
  \caption{In the top-k method, the effect of k value on experimental results. We let the k-max pooling layers have values k is one of 1, 2, 3, 4 and 5. The number of samples is 750, 1500 or 3000 respectively.}\label{fig:experiment1}
\end{figure}

\subsubsection{Accuracy Analysis}

In this experiment, the classification results of the three groups of experiments are evaluated from the three evaluation indexes of accuracy, recall rate and f1-measure[25]. The specific experimental results are shown in Table 3. It can be concluded from Table 3 that the accuracy, recall and f1-score of CNN is 67.33\%, 64.79\% and 67.20\% respectively. Compared with tf-idf and Bigram, the accuracy, recall and f1-measure of CNN have been improved.

\begin{table}[htbp]
  \caption{Effect of OpCNN}
   \label{extremos45}
  \begin{tabular*}{\hsize}{lccc}
 \hline
Experimental method & Accuracy & Recall & F1-measure \\ 
\hline 
tf-idf+svm & 64.53\% & 63.18\% & 64.42\% 
\cr
Bigram+svm & 66.27\% & 64.13\% & 65.85\% 
\cr
CNN & 67.33\% & 64.79\% & 67.20\% 
\cr
OpCNN & 70.02\% & 66.83\% & 69.76\% \\ 
\hline
  \end{tabular*}
\end{table}

Compared with tf-idf+svm, Bigram on the division of the word takes into account the problem of word order to a certain extent. On the other hand, CNN can explore the characteristics of higher latitudes and can reduce the impact of sparseness of data, making the text classification better. If we incorporate CNN model with word order characteristics, we will obtain a more accurate detection result.

Due to the Chinese word order be taken into account the important role of deceptive opinions detection, the k-max pooling method is used to improve the traditional CNN in the pooling layer, which is more suitable for the research of text classification. Through the above experiment, we have set the k value. The experimental group uses the OPCNN model, and the parameters are consistent with CNN. The results of the classification of CNN and OPCNN models are evaluated from the accuracy, the recall and the f1-score. The specific experimental results are shown in Table 3.

From Table 3, compared with 67.33\%, 64.69\% and 67.20\% of CNN, the method has achieved 70.02\%, 66.83\% and 69.76\% of accuracy, recall and f1-measure respectively. In the field of Chinese text categorization, compared with the max pooling method used by the pooling layer, the word order preserving k-max pooling method solves the order problem of Chinese text to some extent. As mentioned earlier in this paper, The classification of the text effect is more obvious.
\subsubsection{Scalability Analysis}
In order to validate the generalization capabilities of the proposed method at the beginning of this chapter, this experiment uses the CNN and OPCNN models on Ott proposing data set. The OTT data set includes 1600 opinions which are divided into four types equally, positive truthful(P and T) opinions, positive deceptive(P and D) opinions, negative truthful(N and T) opinions and negative deceptive(N and D) opinions, as shown in Table 4.  We set the same parameters and use the precision, recall rate and f1-score as evaluation indexes. The specific experimental results are shown in Table 5.

% \begin{table}[t]
% \begin{tabular}{c|c|c|c}
% \multicolumn{4}{l}{\small{\textbf{Table 4}}}\\  
% \multicolumn{4}{l}{\small{Generalization Ability Analysis}}\\ 
% \hline %绘制一条水平的线 
% Experimental method & Accuracy & Recall & F1-score \\ 
% \hline 
% CNN & 82.04\% & 78.20\% & 80.07\% \\ 
% \hline 
% OpCNN & 82.96\% & 80.24\% & 81.58\% \\ 
% \hline
% \end{tabular} 
% \end{table} 

\begin{table}[h]
  \caption{The type of the OTT data set}
   \label{extremos45}
  \begin{tabular*}{\hsize}{lcccc}
\hline %绘制一条水平的线 
Opinion type & P and T & P and D & N and T & N and D\\ 
\hline 
Number & 400 & 400 & 400 & 400 \\ 

\hline
  \end{tabular*}
\end{table}

\begin{table}[h]
  \caption{Generalization Ability Analysis}
   \label{extremos45}
  \begin{tabular*}{\hsize}{lccc}
\hline %绘制一条水平的线 
Experimental method & Accuracy & Recall & F1-measure \\ 
\hline 
CNN & 82.04\% & 78.20\% & 80.07\% \\ 

OpCNN & 84.50\% & 81.03\% & 82.84\% \\ 
\hline
  \end{tabular*}
\end{table}

It can be seen from the experimental results that, on the data set proposed by Ott et al., OPCNN and CNN have the same improvement in the evaluation index. So it can be verified that the method has a good generalization ability. Compared with CNN, the improvement of the evaluation index in this paper is not due to the fact that the method has a certain dependency relationship with the data set used in this paper.

\subsection{Effect of Sample Size}

In order to fully verify the performance advantage of OPCNN compared with other classification methods in deceptive opinions detection, we can compare the classification results by changing the size of the training set. The evaluation index is the accuracy gain($\alpha$). At the same time, in order to prevent the imbalance of the data in the experiment probably having the impact on the experimental results, this experiment uses the same number of deceptive opinion and real opinion. The effect of the number of specific training set samples on the experimental results is shown in Fig.6.

% \begin{figure}[t]
% \setlength{\abovecaptionskip}{0.cm}
% \setlength{\belowcaptionskip}{-2.cm}
% \centering
% \includegraphics[scale=0.4]{accuracy_gain.eps}
% \caption{Accuracy gain($\alpha$). The number of training samples is 250, 500, 1000, 2000, 3000, 4000 respectively.}\label{fig:experiment2}
% \end{figure}

\begin{figure}
\centering
  \includegraphics[width=12cm]{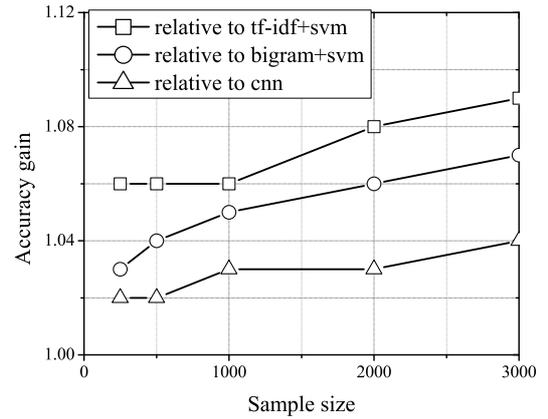}\\
  \vspace{-60pt}
  \caption{Accuracy gain($\alpha$). The number of training samples is 250, 500, 1000, 2000 or 3000 respectively.}\label{fig:experiment2}
\end{figure}

Compared with other methods, the classification method used in this paper obtains the value of $\alpha$ more than 1. At the same time, as the number of samples increases, the accuracy rate of OPCNN model is increasing compared with the other three groups of control experiments, as shown in Fig.6. Since OPCNN and CNN are data driven, with the training sample increasing, the deeper the ability to characterize the depth model has, the higher the accuracy rate is. Compared with CNN, OPCNN has solved the influence of word order on Chinese text classification to a certain extent, so its accuracy is higher. When the number of samples reaches 3000, the accuracy rate is gradually stable, indicating that the accuracy of OPCNN model classification tends to be stable.

%% file: 5analysis.tex
\section{Conclusion}
\label{analysis}
In our paper, the CNN in the deep learning model is used to identify the detective opinions. Against the short opinion text and the various forms of characteristics, we introduce the text order into the deceptive opinion analysis process and extend the scope of the opinion feature. In order to effectively excavate and merge the feature of the opinion text, this paper proposes a guaranteed k-max pooling operation on the basis of CNN. The text order feature is preserved in the process of text feature mining using CNN and the depth of opinion feature is optimized. Experiments show that the improvement of CNN model proposed in this paper can improve the recognition effect of deceptive opinions detection. However, there are still some shortcomings in this paper, such as:  hand-annotated method costs much of manpower. Due to the subjective, the artificial marked data may be awareness of each person to some deviation. In the future experiments, we will continue to improve the above deficiencies to make a better accuracy of deceptive opinions detection.